\def\year{2022}\relax
\documentclass[letterpaper]{article} 
\usepackage{aaai22}  
\usepackage{times}  
\usepackage{helvet}  
\usepackage{courier}  
\usepackage[hyphens]{url}  
\usepackage{graphicx} 
\usepackage{natbib}  
\usepackage{caption} 
\DeclareCaptionStyle{ruled}{labelfont=normalfont,labelsep=colon,strut=off} 
\usepackage{algorithm}
\usepackage{algorithmic}

\usepackage{amsmath}
\usepackage{amssymb}
\usepackage{booktabs}
\usepackage{bm}

\usepackage{newfloat}
\usepackage{listings}
\floatstyle{ruled}
\newfloat{listing}{tb}{lst}{}
\floatname{listing}{Listing}
\title{Contrast and Generation Make BART a Good Dialogue Emotion Recognizer}
\author{
    Shimin Li\textsuperscript{\rm 1,3},
    Hang Yan\textsuperscript{\rm 1,3},
    Xipeng Qiu\textsuperscript{\rm 1,2,3}\thanks{Corresponding Author.}
}
\affiliations{
    \textsuperscript{\rm 1} School of Computer Science, Fudan University\\
    \textsuperscript{\rm 2} Peng Cheng Laboratory, Shenzhen, Guangdong, China\\
    \textsuperscript{\rm 3} Shanghai Key Laboratory of Intelligent Information Processing, Fudan University

    \{smli20, hyan19, xpqiu\}@fudan.edu.cn
}

\begin{document}

\maketitle

\begin{abstract}
In dialogue systems, utterances with similar semantics may have distinctive emotions under different contexts. Therefore, modeling long-range contextual emotional relationships with speaker dependency plays a crucial part in dialogue emotion recognition. Meanwhile, distinguishing the different emotion categories is non-trivial since they usually have semantically similar sentiments. To this end, we adopt supervised contrastive learning to make different emotions mutually exclusive to identify similar emotions better. Meanwhile, we utilize an auxiliary response generation task to enhance the model's ability of handling context information, thereby forcing the model to recognize emotions with similar semantics in diverse contexts. To achieve these objectives, we use the pre-trained encoder-decoder model BART as our backbone model since it is very suitable for both understanding and generation tasks. The experiments on four datasets demonstrate that our proposed model obtains significantly more favorable results than the state-of-the-art model in dialogue emotion recognition. The ablation study further demonstrates the effectiveness of supervised contrastive loss and generative loss\footnote{\url{https://github.com/whatissimondoing/CoG-BART}.}.

\end{abstract}

\section{Introduction}
\begin{figure}[t]
\centering
\includegraphics[width=0.9\columnwidth]{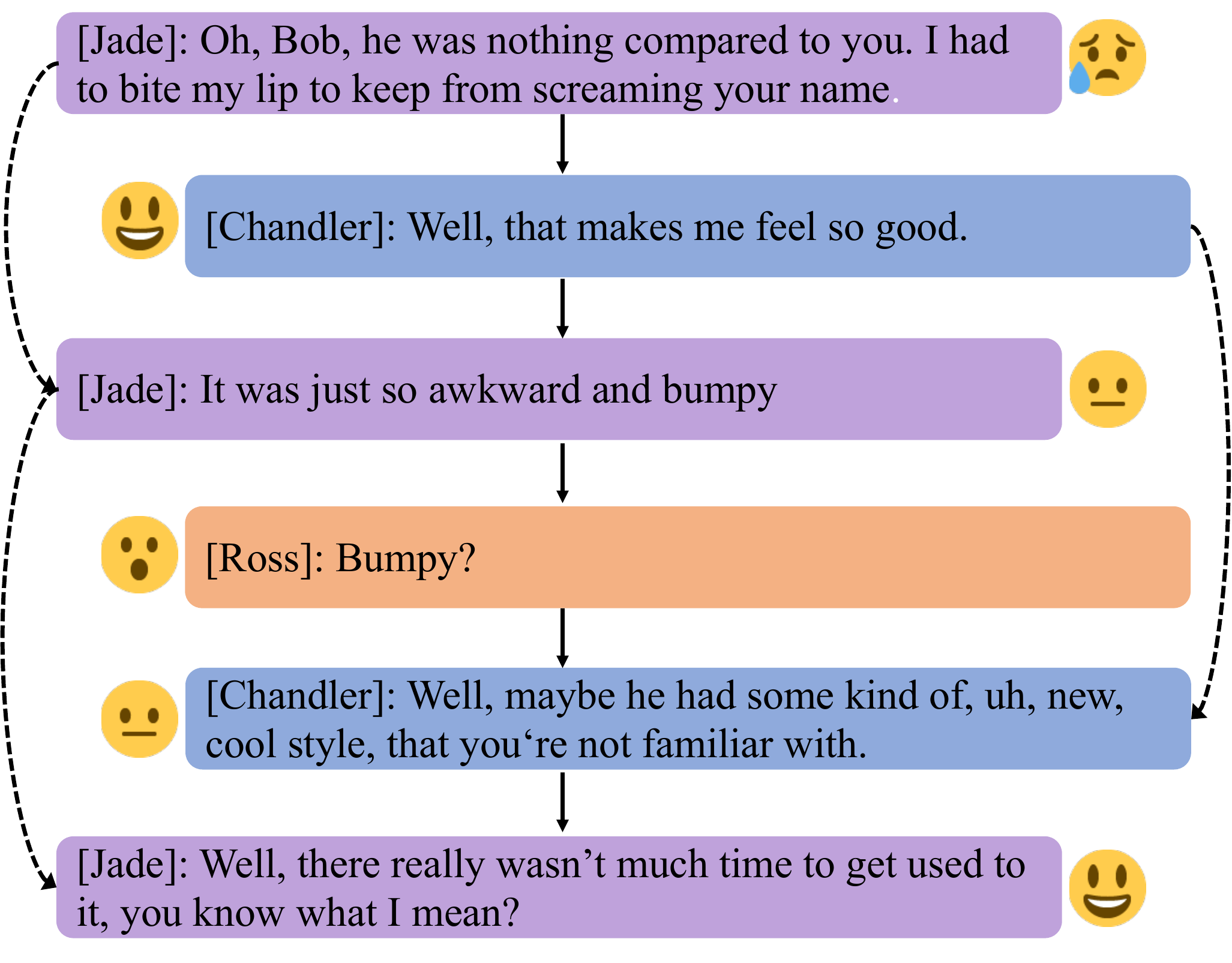} 
\caption{The conversation flow chart in multi-person dialogue emotion recognition. The solid line indicates that the previous utterance directly influences the current speaker's emotion. The dashed line signifies that the same speaker is influenced by other utterances and expresses different emotions.}
\label{task_intro}
\end{figure}

With the development and popularization of personal intelligent terminal technology and social networks, the importance of constructing a dialogue system that can comprehend user emotions and intentions and conduct effective dialogue interactions has increased significantly. A critical module in the dialogue system is the natural language understanding module that analyzes user behaviors like intents or emotions. Analyzing user sentiments with contextual relationships is an advanced step for simple sentiment classification tasks and is more suitable for usage scenarios in the real world with more research value. The task of emotion recognition in conversation (ERC) is to assign emotion labels to all the utterances in a historical dialogue with a contextual relationship. At the same time, each historical dialogue contains interactions between multiple different speakers, as illustrated in Figure \ref{task_intro}.

There are three challenges for ERC. (1) The first challenge is that the emotion of each utterance may be affected by contextual information. For example, specific emotions will depend on certain utterances of the context. Meanwhile, utterances with the same expression may have completely different emotions in various contexts. Therefore, effectively modeling the context dependency and the speaker dependency is the main factor distinguishing this task from traditional sentiment classification. (2) The second challenge is that each speaker's emotion is influenced by the utterance of other speakers in the conversation, so there may exist a sudden change in a speaker's emotion. (3) The third challenge lies in semantically similar but different categories of emotions, such as ``frustrated'' to ``sad'', ``happy'' to ``excited'', etc. It is difficult to distinguish these semantically similar sentiment categories.

Recent related work addressed contextual dependencies and speaker relations using various graph networks \cite{dag-erc, ghosal-etal-2019-dialoguegcn, rgat, sheng-etal-2020-sumagggin}. However, as the number of layers deepens, the phenomenon of over-smoothing \cite{over-smoothing-aaai-20} starts to appear, resulting in the representation of similar sentiments tending to be indistinguishable.

This work deals with the above challenges by better modeling the context and speaker information and auxiliary generation task.

Firstly, we introduce a dialogue-level Transformer \cite{transformer-nips-2017} layer to model the long-range context dependencies between utterances. A pre-trained language model captures the representation of each utterance. Compared to previous approaches that only adopt pre-trained models as a feature extractor \cite{roberta} and employ the extracted features as the node representation of downstream graph networks, a pure Transformer structure makes fewer prior structural assumptions \cite{transformer_survey}.

Secondly, we adopt supervised contrastive learning (SCL) \cite{khosla2020supervisedcl} to alleviate the difficulty in categorizing similar emotions, which makes samples with same sentiments cohesive and different sentiments mutually exclusive under the fully utilization of label information. Compared with the cross-entropy loss for noisy labels, the supervised contrastive loss can increase the stability of training and improve the generalization of the model \cite{scl-fine-tune}. 
Unlike the regular SCL, we copy the hidden state of all samples in a batch and detach off its gradient as its multiview representation. The reason is that the categories in existing ERC datasets are highly unbalanced, and some categories may exist in a batch with only one sample. If only the original SCL is used, it will lead to incorrect loss calculation.

Thirdly, we introduce an auxiliary response  generation task to enhance the ability of capturing the context information for ERC. The prediction of the following utterance makes the model fully consider contextual dependencies, thus forcing the model to consider the information in the context and rely on the current utterance itself when recognizing the sentiment in the conversation. Moreover, by splicing the speaker directly before utterance as a hint for speaker information, the dependency between speakers and utterances is modeled adequately without additional parameters.

Finally, we utilize BART \cite{lewis-etal-2020-bart}, a pre-trained Transformer with an encoder-decoder structure, as our backbone model and enhance it by contrastive and generative loss. Our proposed \textbf{CO}nstrastive-and-\textbf{G}eneration-enhanced BART (CoG-BART) obtains state-of-the-art results on four ERC datasets compared to the baseline models. Additionally, ablation experiments and case studies prove the effectiveness of the contrastive and generative losses in the ERC task.

To summarize, our main contributions can be concluded as follows:
\begin{itemize}
\item To the best of our knowledge, we utilize supervised contrastive learning for the first time in ERC and significantly improve the model's ability to distinguish different sentiments.
\item By incorporating response generation as an auxiliary task, the performance of ERC is improved when certain contextual information is involved.
\item Our model is easy-to-implemented since it does not depend on external resources, like graph-based methods.
\end{itemize}

\begin{figure*}[t]
\centering
\includegraphics[width=0.95\textwidth]{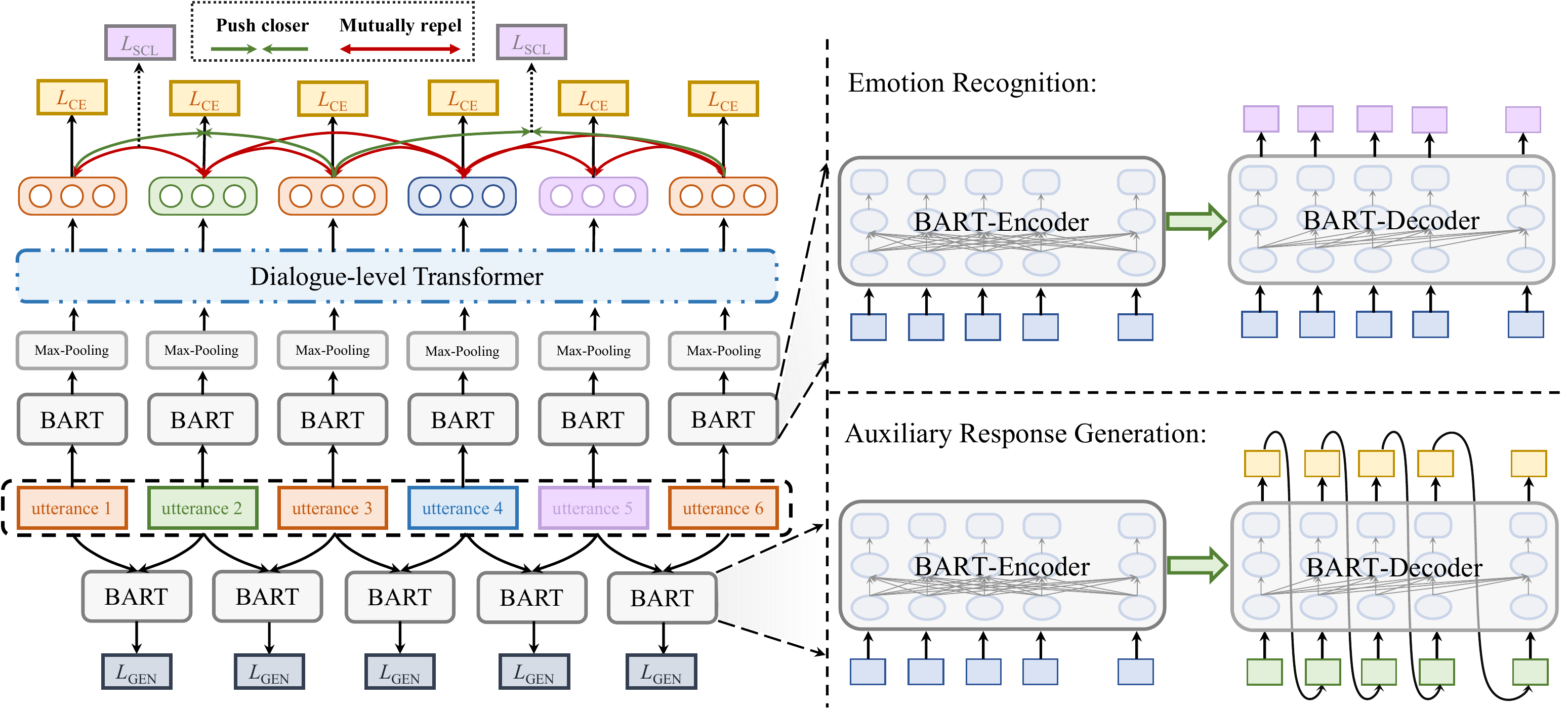} 
\caption{The overall framework of CoG-BART. The utterance is fed into BART for $N$ utterances in a batch to get its hidden state. The representation of the utterance obtained after max-pooling the hidden state of each utterance is fed to the upper-level dialogue-level Transformer for modeling context dependencies. The obtained context-dependent utterance representations are utilized to compute the cross-entropy loss and supervised contrastive loss. In addition, the two adjacent utterance pairs are used for the auxiliary response generation.}
\label{fig1}
\end{figure*}

\section{Related Work}
This section will introduce related works in ERC. Due to context-dependency and speaker dependency properties, it is natural for researchers to employ graph neural networks. Therefore, many works have constructed various task-specific graph structures. Meanwhile, with the excellent performance of the pre-trained model in diverse downstream tasks, an increasing number of works adopt the pre-trained model as the feature extractor for the input of the downstream model or directly fine-tune it with downstream datasets. Therefore, this section divides the related work into two categories: graph-based models and pre-train-based models.

\subsection{Dialog Emotion Recognition}
\subsubsection{Graph-based model}
Considering the unidirectionality of information interaction, DAG \cite{dag-erc} utilizes directed acyclic graphs to model the information interaction between utterance and speaker. DialogGCN \cite{ghosal-etal-2019-dialoguegcn} adopts the basic graph neural network to model the relationship between contexts. SumAggGIN \cite{sheng-etal-2020-sumagggin} adds an aggregation module based on DialogGCN to additionally consider phrase-level information other than utterance-level. By simulating the process of human reasoning, DialogCRN \cite{dialogcrn} proposes to apply several reasoning modules to extract and integrate clues of emotional reasoning. To make the model better understand the additional general information involved in the dialogue process, KET \cite{zhong-2019-ket} combines external knowledge with a hierarchical Transformer. By appending sequence information into the graph network, RGAT \cite{rgat} uses relational position encoding to combine position information into the graph network structure to consider the dependency between speakers. TODKAT \cite{todkat} integrates topic detection into the pre-training model and fuses commonsense knowledge into Transformer \cite{transformer-nips-2017}.

\subsubsection{Pre-train-based model}
Suppose each utterance is regarded as an independent sentence, regardless of its context-dependence and speaker information. In that case, the problem can be transformed into a simple sentence classification so that pre-trained models \cite{ptm_survey} such as BERT \cite{devlin-etal-2019-bert}, BART \cite{lewis-etal-2020-bart}, and RoBERTa \cite{roberta} can be used directly for fine-tuning. HiTrans \cite{li-etal-2020-hitrans} adopts BERT to extract utterance features, followed by transformer structure for modeling context. Considering speaker dependence, the auxiliary task of judging whether two utterances are the same speaker is used to model the speaker information. COSMIC \cite{ghosal-etal-2020-cosmic} exploits RoBERTa as the feature extractor of each utterance and model the dependency of the context with RNN. In addition, the common knowledge transformer COMET \cite{bosse-20-comet} is incorporated to introduce world knowledge. Based on XLNet, DialogXL \cite{dialog-xl} changes the segment-level structure to utterance-level and uses memory to record the historical context. Meanwhile, by adopting different mask mechanisms on different attention heads, each attention head pays attention to different aspects of dialogue information. \citet{gen-and-cls-naacl} trained BART with both generation and classification in a multi-task format, though their method focused on response generation, treating emotion recognition as an auxiliary task. However, we focus on ERC and apply supervised contrastive loss as an additional optimization goal.

\subsection{Contrastive Learning}
\subsubsection{Unsupervised contrastive learning}
In the field of computer vision, SimCLR \cite{ChenK0H20-simclr} takes pictures obtained from the same image through randomly different data augmentation methods as positive samples and other pictures as negative samples, thereby optimizing contrastive loss. The naive sentence representation obtained by BERT has poor performance in semantic text similarity tasks. Therefore, ConSERT \cite{YanLWZWX20-consert} introduces self-supervised contrast loss in the fine-tuning stage of BERT. MBERT \cite{kim-21-mbert} does not use data augmentation to construct positive samples but uses BERT with frozen parameters and fine-tunable parameters as a special siamese model to construct positive samples.
\subsubsection{Supervised contrastive learning}
To make full use of label information, \citet{khosla2020supervisedcl} extends it to supervised contrastive learning based on self-supervised training so that samples belonging to the same label are gathered in the embedding space while pushing samples of different categories away. Given that cross-entropy loss may cause model training instability and converge to a local optimum, SCL \cite{scl-fine-tune} introduces supervised contrastive loss in the fine-tuning stage, which greatly improves the model's performance in few-shot learning scenarios. SimCSE \cite{simcse} uses entailment pair in the annotated NLI dataset as the positive sample and the contradict pair as the negative sample in supervised contrastive learning.

\section{Methodology}
\subsection{Problem Definition}
In dialogue emotion recognition, the data is composed of multiple conversations $ \{c_1, c_2, \cdots, c_N\} $, with each conversation composed of several utterances $ c_i= \left[ u_1, u_2, \cdots, u_m \right] $ and emotion labels $ \mathcal{Y}_{c_i}=\{y_1, y_2, \cdots, y_m\}\in S$, where $S$ indicates the categories of emotions. For an utterance, it is comprised of several tokens $ u_t=\left[w_{t,1},w_{t,2}, \cdots, w_{t,n}\right] $. Every utterance in a conversation $c_i$ is uttered by one speaker which can be represented as $ p(c_i)=\left[p(u_1),\cdots,p(u_i),\cdots,p(u_m)\right]$ and $p(u_i)\in P $, where $P$ indicates the categories or names of the speakers. Accordingly, the whole problem can be expressed as getting the emotional label of each utterance based on the context and speaker information in a piece of conversation: $ \mathcal{Y}_{c_i}=f(c_i, p(c_i)) $.

\subsection{Supervised Contrastive Learning for ERC}
\subsubsection{Utterance Encoding}
To model the dependencies between speaker and utterance, for a certain utterance $u_{t}$ in a conversation, we splice the speaker's name or category before the utterance.
After tokenizing the utterance prepended with the speaker information, we get:
\begin{flalign}
    & \tilde{u}_{t}=\left[\langle s\rangle,w_{t,1},\cdots, w_{t,i},\cdots, w_{t,|n_{t}|}, \langle /s\rangle\right],
\end{flalign}

\noindent where $\langle s\rangle$ and $\langle /s\rangle$ are treated as special tokens to indicate the beginning and end of an utterance.
Then the token sequence after tokenization is fed to the shared embedding layer of BART to acquire the hidden state of each token in utterance before sending it to the encoder and decoder of BART. After sending $H_t$ to BART, the representation of the current utterance $\widehat{H}_{t}$ is acquired:
\begin{flalign}
     H_{t}&=\mathrm{EmbeddingLayer}(\tilde{u}_{t}), \\
     \widehat{H}_{t}&=\mathrm{BART\textrm{-}Model}(H_{t}),
\end{flalign}

\noindent where $H_t,\widehat{H}_t \in\mathbb{R}^{s\times d}$, and $s, d$ indicates the length of the sequence and hidden dimension respectively.

\subsubsection{Dialogue Modeling}

The representation $\widehat{H}_{t}$ obtained by the BART-Model is max-pooled to obtain the aggregated representation of the utterances as follows:
\begin{flalign}
    & \check{h}_{t}=\mathrm{max\textrm{-}pooling}(\widehat{H}_{t}).
\end{flalign}


To model the historical context information of the dialogue, we exploit a dialogue-level Transformer \cite{transformer-nips-2017} layer as the context encoder. The multi-head attention mechanism can capture the interaction between different dialogues in multiple rounds of dialogue and aggregate different features to obtain the final implicit representation, thereby fully modeling the complex dependence between different utterances and context relations. For all utterances in a context, the multi-head attention score of the hidden state between two different utterances in a conversation $\check{h}_{j}$, $\check{h}_{k}$ can be calculated by the following formulas:

\begin{flalign}
    &\mathrm{Atten}(Q,K,V)=\mathrm{softmax}(\frac{QK^T}{\sqrt{d_k}})V, \\
    &head_i=\mathrm{Atten}(\check{h}_{j}W_i^Q, \check{h}_{k}W_i^K,\check{h}_{k}W_i^V), \\
    &\mathrm{MultiHead}(Q,K,V)=[head_1;\cdots; head_n]W^O,
\end{flalign}

\noindent where $W_i^Q\in \mathbb{R}^{d\times d_q}$, $W_i^K\in \mathbb{R}^{d\times d_k}$, $W_i^V\in \mathbb{R}^{d\times d_v}$ and $ W^O\in \mathbb{R}^{d\times d}$ are parameters that can be optimized, $d_q$, $d_k$ and $d_v$ are dimensions of query, key and value vectors, $n$ indicates the number of heads.

Therefore, the utterance representation that models the context-dependence can be obtained through the above-mentioned dialogue-level Transformer:
\begin{flalign}
    {H}_{win}&=[\check{h}_t, \check{h}_{t+1}, \cdots, \check{h}_{t+bs-1}], \\
    {H}_{d\textrm{-}win}&=\mathrm{Dialogue}\textrm{-}\mathrm{Transformer}({H}_{win}),
\end{flalign}

\noindent where ${H}_{win}\in\mathbb{R}^{bs\times d}$ indicates utterances in a conversation within the window size $bs$ and ${H}_{d\textrm{-}win}\in\mathbb{R}^{bs\times d}$ denotes the utterances after context modeling.

\subsubsection{Supervised Contrastive Learning}

Supervised contrastive learning assumes that some crucial aspects get attention and allows few-shot learning to be more stable when fine-tuned on pre-trained models \cite{scl-fine-tune}.
The typical contrastive learning uses only one pair of positive examples, while all other samples are treated as negative examples. Supervised contrastive learning treats all examples with the same label in the batch as positive examples by making full use of label information.




For ERC, the number of samples in each category in some datasets \cite{li-etal-2017-dailydialog} is highly unbalanced, while the supervised contrastive learning will mask itself when calculating the loss. If only one sample exists for a category in the batch, it cannot be directly applied to calculate the loss. Therefore, a copy of the hidden state of the utterance ${H}_{d\textrm{-}win}$ is made to obtain $\overline{H}_{d\textrm{-}win}$, and its gradient is detached. Hence the parameter optimization is maintained stable.

For a batch with $N$ training samples, each sample is operated by the above mechanism to obtain multiview $2N$ samples, then the supervised contrastive loss of all samples in a batch can be expressed by the following equation:

\begin{flalign}
    X&=[{H}_{d\textrm{-}win},\overline{H}_{d\textrm{-}win}], \\
    \mathcal{L}_{\mathrm{SCL}}&=\sum_{i\in I}\frac{-1}{|P(i)|}\sum_{p\in P(i)}\textrm{SIM}(p,i), \\
    \textrm{SIM}(p,i)&=\log\frac{\exp((X_i\cdot X_p)/\tau)}{\sum_{a\in A(i)}\exp(X_i\cdot X_a/\tau)},
\end{flalign}

\noindent where $X\in \mathbb{R}^{2N\times d}$, $i\in I=\{1,2,\cdots,2N\}$ indicate the index of the samples in a multiview batch, $\tau\in R^+$ denotes the temperature coefficient used to control the distance between instances, $P(i)=I_{j=i}-\{i\}$ represents samples with the same category as $i$ while excluding itself, $A(i)=I-\{i, N+i\}$ indicates samples in the multiview batch except itself.

\subsection{Auxiliary Response Generation}
To facilitate the model to consider richer contextual information when determining utterance sentiment, the model is required to generate its following utterance $u_{t+1}$ given the current utterance $u_t$. The output hidden state of each token in $u_{t+1}$ is generated by the BART decoder sequentially.

\begin{flalign}
     \acute{H}_t&=\textrm{BART-Encoder}(H_t), \\
     \grave{h}_j^d&=\textrm{BART-Decoder}(\acute{H}_t;\grave{h}_{<j}^d), \\
     u_{t+1,j}&=\textrm{Softmax}(\grave{h}_j^d), \\
     \mathcal{L}_{\mathrm{Gen}}&=-\sum_{i=1}^N\log p(u_{t+1}|u_t, \bm{\theta}),
\end{flalign}
\noindent where $\bm{\theta}$ is the parameters of BART need to be optimized.

\subsection{Model Training}
The loss of model training consists of three parts: the hidden state ${H}_{d\textrm{-}win}$ obtained after context modeling passes through a multilayer perceptron to obtain logits for calculating cross-entropy loss. The other part is the supervised contrastive loss and the loss of response generation. The loss is a weighted sum of the three components, and the sum of their weights equals one. The overall framework of CoG-BART is illustrated in Figure \ref{fig1}.

\begin{flalign}
    P_i&=\mathrm{Softmax}(W_s {H}_{d\textrm{-}win,i}+b_s), \\
    \hat{y}_i&=\mathrm{argmax}(P_i), \\
    L_{\textrm{CE}}&=-\frac{1}{N}\sum_{i=1}^N\sum_{c=1}^Cy_{i,c}\cdot \log \hat{y}_{i,c}, \\
    \mathcal{L}&=(1-\alpha-\beta)\mathcal{L}_{\mathrm{CE}}+\alpha \mathcal{L}_{\mathrm{SCL}}+\beta\mathcal{L}_{\mathrm{Gen}},
\end{flalign}

\noindent where $y_{i,c}$ represents the label of a certain utterance, $\hat{y}_{i,c}$ indicates the probability distribution of category $c$ output by the dense layer, $\alpha$ denotes the weight for supervised contrastive loss and $\beta$ is the weight for loss of response generation.

\begin{table}[t]
\centering
\resizebox{.95\columnwidth}{!}{

\begin{tabular}{clcccc}
\toprule
{Dataset} &       & DD    & MELD & ENLP & IEMOCAP \\
\midrule
{\#Dial}  & Train & 11118 & 1038 & 713  & 120     \\
{}        & Dev   & 1000  & 114  & 99   & 120     \\
{}        & Test  & 1000  & 280  & 85   & 31      \\
{\#Utter} & Train & 87170 & 9989 & 9934 & 5810    \\
{}        & Dev   & 8069  & 1109 & 1344 & 5810    \\
{}        & Test  & 7740  & 2610 & 1328 & 1623    \\
\midrule
{\#CLS}   &       & 7     & 7    & 7    & 6       \\
\bottomrule
\end{tabular}
}
\caption{Statistics of four benchmark datasets. }
\label{dataset_desc}
\end{table}

\begin{table*}[t]
\centering
\resizebox{\textwidth}{!}{
\setlength{\tabcolsep}{.65mm}
\begin{tabular}{lcccccccc}
\toprule
Dataset      & \multicolumn{2}{c}{MELD}                                   & \multicolumn{2}{c}{EmoryNLP}
             & \multicolumn{2}{c}{IEMOCAP}                                & \multicolumn{2}{c}{DailyDialog}      \\
\midrule
Model        & Weighted           & Micro-F1                              & Weighted          & Micro-F1
             & Weighted           & Micro-F1                              & Weighted          & Micro \\
{}           & -Avg-F1            &                                       & -Avg-F1           &
             & -Avg-F1            &                                       & -F1-neural   & -F1-neutral \\
\midrule
BERT         & 62.28              & 63.49                                 & 34.87             & 41.11
             & 60.98              & -                                     & 53.41             & 54.85             \\
RoBERTa      & 62.51              & 63.75                                 & 35.90             & 40.81
             & 63.38              & -                                     & 52.84             & 54.33             \\
HiTrans      & 61.94              & -                                     & 36.75             & -
             & 64.50              & -                                     & -                 & -                 \\
DialogXL     & 62.41              & -                                     & 34.73             & -
             & 65.94              & -                                     & -                 & 54.93             \\
XLNet        & 61.65              & -                                     & 34.13             & -
             & 61.33              & -                                     & -                 & 53.62             \\
BART-large   & 63.57              & 64.41                                 & 35.98             & 38.93
             & 56.14              & 56.67                                 & 54.83             & 55.34            \\
\midrule

CoG-BART     & \textbf{64.81} ($\pm$0.19)       & \textbf{65.95} ($\pm$0.44) & \textbf{39.04} ($\pm$0.10)      & \textbf{42.58} ($\pm$0.94) & \textbf{66.18} ($\pm$0.45)                  & \textbf{66.71} ($\pm$0.49)            & \textbf{56.09} ($\pm$0.01)             & \textbf{56.29} ($\pm$0.17)                  \\
\bottomrule
\end{tabular}

}
\caption{The overall results of CoG-BART with pre-train-based baseline models on four datasets. }
\label{main_results}
\end{table*}

\begin{table}[t]
\centering
\resizebox{1\columnwidth}{!}{
\setlength{\tabcolsep}{.3mm}
\begin{tabular}{@{}lcccc@{}}
\toprule
Dataset      & \multicolumn{1}{c}{MELD}                                   & \multicolumn{1}{c}{EmoryNLP}
             & \multicolumn{1}{c}{IEMOCAP}                                & \multicolumn{1}{c}{DailyDialog}      \\
\midrule
Model        & Weighted                                                   & Weighted
             & Weighted                                                   & Micro            \\
{}           & -Avg-F1                                                    & -Avg-F1
             & -Avg-F1                                                    & -F1-neutral      \\
\midrule
KET          & 58.18                                                      & 34.39
             & 59.56                                                      & 53.37            \\
RGAT         & 60.91                                                      & 34.42
             & 65.22                                                      & 54.31            \\
RGAT+RoBERTa & 62.80                                                      & 37.89
             & 66.36                                                      & 59.02            \\
DialogGCN    & 58.10                                                      & -
             & 64.18                                                      & -                \\
DialogCRN    & 58.39                                                      & -
             & 66.20                                                      & -                \\
COSMIC       & 64.28                                                      & 37.10
             & 63.05                                                      & 56.16            \\
DAG-ERC      & 63.65                                                      & 39.02
             & \textbf{68.03}                                             & \textbf{59.33}   \\
\midrule
CoG-BART     & \textbf{64.81} ($\pm$0.19)                                 & \textbf{39.04} ($\pm$0.10)
             & 66.18 ($\pm$0.45)                                          & 56.29 ($\pm$0.17)                  \\

\bottomrule
\end{tabular}
}
\caption{Comparison with graph-based models.}
\label{main_results_graph_model}
\end{table}

\section{Experimental Settings}
This section will elaborate on the datasets, baseline models, experimental conditions, and parameter settings adopt in the experiment.
\subsection{Experimental Setup}
The code framework and initial weight of BART come from Huggingface's Transformers \cite{wolf-etal-2020-transformers}. The optimizer applied for model training is AdamW with a linear-scheduled warm-up strategy. The parameters adjusted in this experiment include batch size, learning rate, warm-up ratio, $\alpha$, and $\beta$. We conducted a hyperparameter search for model training through the reserved validation set. The results on the test set come from the best checkpoint in the validation set, and we average the scores from five different random seeds. All experiments are performed on GeForce RTX 3090 GPU.

\subsection{Datasets}
This section will introduce four benchmark datasets: MELD \cite{poria-etal-2019-meld}, EmoryNLP \cite{emorynlp}, DailyDialog \cite{li-etal-2017-dailydialog}, and IEMOCAP \cite{busso2008iemocap} for comparison with the baseline models.
\subsubsection{MELD}
This dataset comes from the dialogue content of the characters in the American drama \textit{Friends}. MELD originally contained multi-modal data, but we used only the text data for the experiments.
\subsubsection{EmoryNLP (ENLP)}
This dataset also comes from \textit{Friends}, and the difference from MELD is the annotation of utterance's emotional label category. The emotional tags contained in this dataset are: \textit{joyful}, \textit{neutral}, \textit{powerful}, \textit{mad}, \textit{sad}, \textit{scared}, and \textit{peaceful}.
\subsubsection{DailyDialog (DD)}
Manually compiled data sets about daily communication. The annotation method used in this data set is Ekman's emotion type \cite{ekman1993facial}, which includes six basic emotion tags, including \textit{happiness}, \textit{surprise}, \textit{anger}, \textit{disgust}, \textit{fear}, and \textit{sadness}.
\subsubsection{IEMOCAP}
Like MELD, it is a multi-modal dataset. The content is derived from the lines in the scripts of the two actors, and the emotional tags included are \textit{excited}, \textit{neutral}, \textit{frustrated}, \textit{sad}, \textit{happy}, and \textit{angry}.

The detailed statistics of the four datasets are shown in Table \ref{dataset_desc}, where ``\#Dial" indicates the number of dialogue in train/dev/test, ``\#Utter" represents the number of all utterances in dialogue, and ``\#CLS" denotes the number of categories of each dataset.

\subsection{Metrics}
For MELD, EmoryNLP and IEMOCAP, we adopt weighted average F1 as the evaluation metrics. In that ``neutral" occupies the majority in DailyDialog, micro-F1 is employed as the evaluation metric for this data set, and we ignore the label ``neutral" when calculating the results as in the previous works \cite{todkat, dag-erc}.

\section{Results and Analysis}
\subsection{Main Results}

Table \ref{main_results} and \ref{main_results_graph_model} record the results of comparing CoG-BART with the baseline models on four datasets.

Among the pre-train-based models and their variants, the selected baseline models are BERT \cite{devlin-etal-2019-bert}, RoBERTa \cite{roberta}, HiTrans \cite{li-etal-2020-hitrans}, DialogXL \cite{dialog-xl} and XLNet \cite{xlnet}.
In MELD \cite{poria-etal-2019-meld}, CoG-BART has an approximate absolute 1.24\% improvement over the previous state-of-the-art BART-large \cite{lewis-etal-2020-bart}.

For graph-based models, KET \cite{zhong-2019-ket}, RGAT \cite{rgat}, DialogGCN \cite{ghosal-etal-2019-dialoguegcn}, DialogCRN \cite{dialogcrn},  COSMIC \cite{ghosal-etal-2020-cosmic}, and DAG-ERC \cite{dag-erc} are listed.

Compared to the graph-based model, CoG-BART improves 0.53 points over COSMIC \cite{ghosal-etal-2020-cosmic}. It is worth noting that RoBERTa-large is used as the feature extractor in COSMIC, while CoG-BART only adopts BART-large as the backbone structure to obtain competitive results, indicating that adequate knowledge transfer of pre-trained models which  effectively model the dependencies between contexts can also yield promising results in MELD.

We can observe from the results in EmoryNLP \cite{emorynlp} that the graph-based model using the pre-trained model as the feature extractor works better overall than the model applying only the pre-trained model as the backbone network. Meanwhile, CoG-BART still achieves results with significant improvement. Also, the graph-based model can obtain higher F1 overall on IEMOCAP \cite{busso2008iemocap} compared to the pre-trained based models. The reason is that the number of utterances contained in one context of IEMOCAP is much larger than the other three datasets, so pre-trained models are usually incapable of handling excessively long contexts. However, graph network models can better model context dependencies. In comparison, CoG-BART also achieves results similar to those of graph-based models, demonstrating the capability of CoG-BART to model the context-dependence.

The micro-F1 values of CoG-BART in DailyDialog are lower compared to the results of some graph neural network models. Still, it can achieve similar results to some pre-train-based models such as BERT \cite{devlin-etal-2019-bert}, RoBERTa \cite{roberta} and DialogXL \cite{dialog-xl}. Therefore, the graph-based model may have the advantage over pre-train-based models by more adequately modeling context dependencies on this dataset.

\begin{figure}[t]
\centering
\includegraphics[width=0.98\columnwidth]{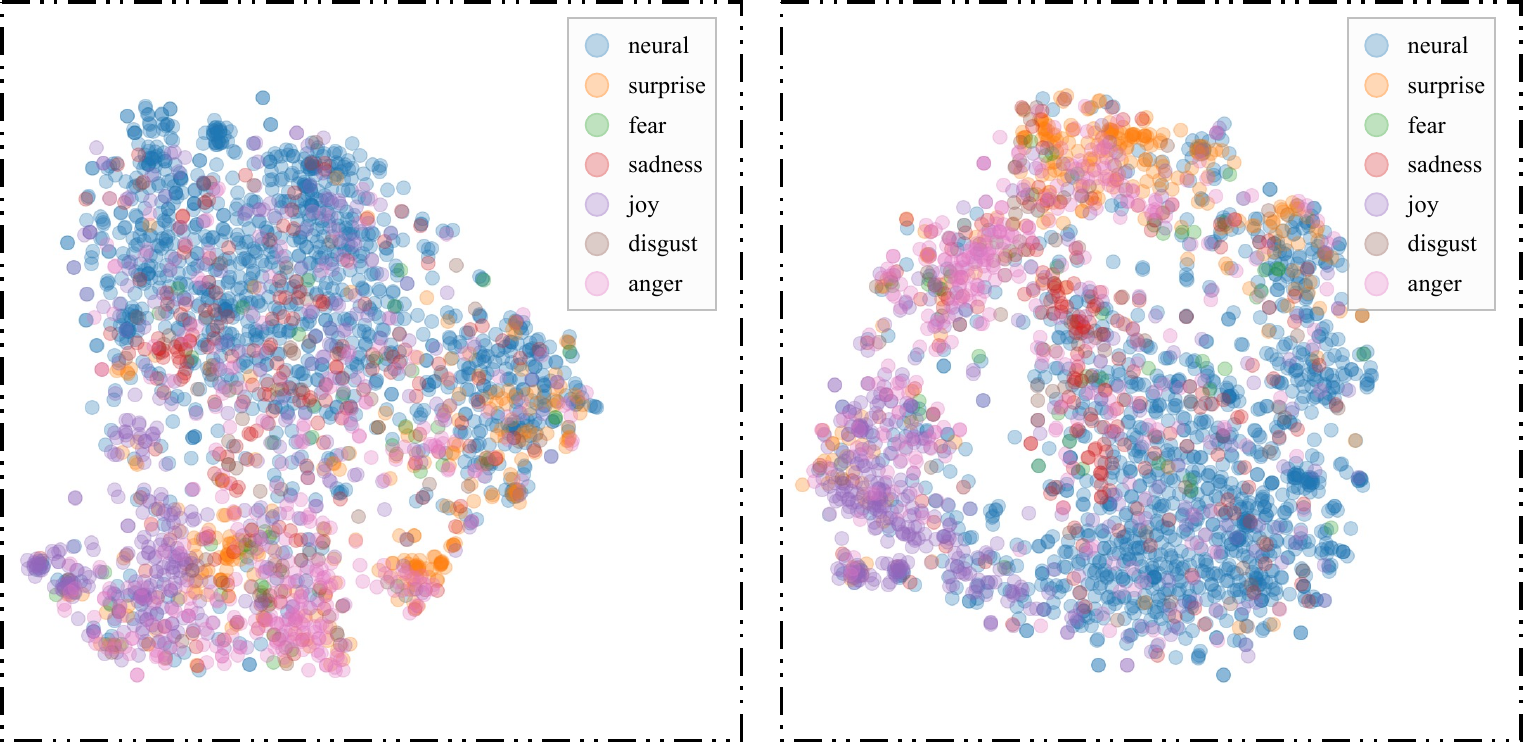} 
\caption{The t-SNE visualization results of the model output when $\alpha$ is 0 and 0.8, respectively.}
\label{effect_of_scl}
\end{figure}

\subsection{The Potency of Supervised Contrastive Learning}
\subsubsection{Qualitative Analysis of SCL}
To conduct a qualitative analysis of supervised contrastive learning, we utilize t-SNE \cite{t-sne-2002} to visualize the distribution of high-dimensional hidden states obtained by the model trained with supervised contrastive loss. By controlling different sizes of $\alpha$, the ratio of supervised contrastive loss is controlled to 0\% and 80\%, respectively, to obtain the hidden state output by the model under different levels of supervised contrastive learning.

As illustrated in Figure \ref{effect_of_scl}, when the supervised contrastive loss is not exploited, that is, when the cross-entropy loss function is completely adopted, the overlap rate of samples between different labels is particularly high, especially for some samples with similar emotions, which increase the difficulty of learning the decision boundaries. As the proportion of supervised contrastive loss increases, it can be distinctly observed that the degree of coupling between different classes is gradually enlarged, and the same classes begin to cohesive. It is worth mentioning that although the distance within the class has been reduced, the uniformity \cite{align_and_uniform} between samples has been well maintained, indicating that the information has been well preserved and no representation collapse has occurred.

\begin{table}[t]
\centering
\resizebox{1\columnwidth}{!}{
\begin{tabular}{lcccccc}
\toprule
{Metric}                  & \multicolumn{6}{c}{Weighted Average F1}                                                             \\
\midrule
{Datasets}                & $\alpha$=0.2    & $\alpha$=0.4   & $\alpha$=0.6 & $\alpha$=0.8     & $\beta$=0.1    & $\beta$=0.2   \\
\midrule
MELD                      & \textbf{64.57}  & 63.99          & 64.42        & 61.84            & 64.83          & 63.70         \\
IEMOCAP                   & 64.38           & \textbf{66.18} & 65.12        & 63.38            & 66.18          & 63.54         \\
EmoryNLP                  & \textbf{39.04}  & 36.68          & 36.90        & 35.24            & 37.45          & 37.57         \\
\bottomrule
\end{tabular}
}
\caption{The F1 scores for different values of $\alpha$ and $\beta$}
\label{alpha_of_scl}
\end{table}

\begin{figure*}[t]
\centering
\includegraphics[width=0.95\textwidth]{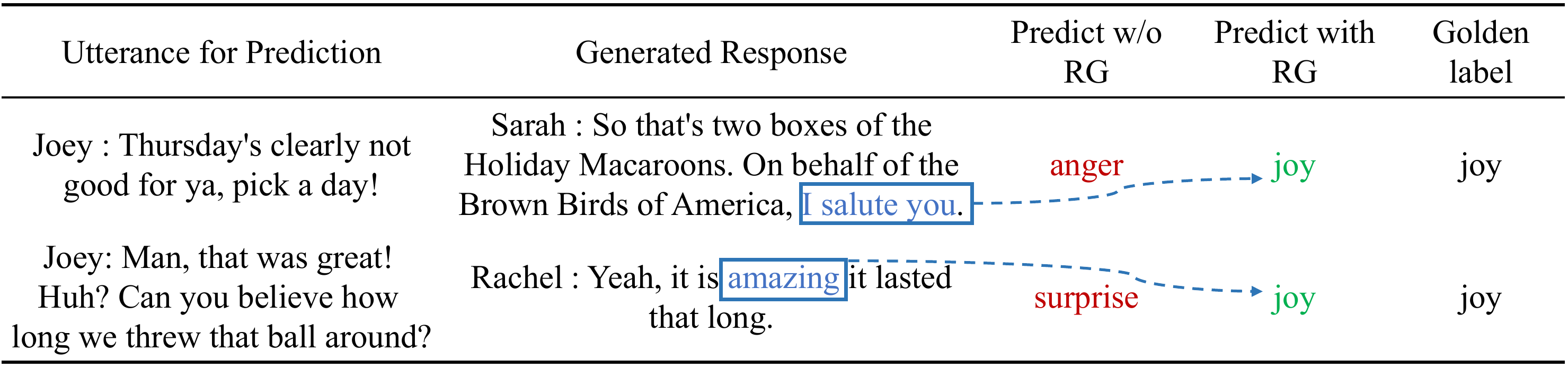} 
\caption{Case studies show that response generation enables the model to correctly predict the emotion based on context.}
\label{effect_of_response_generation}
\end{figure*}

\subsubsection{Quantitative Analysis of SCL}
The effects of different proportions of supervised contrastive learning on CoG-BART are illustrated in Table \ref{alpha_of_scl}, where the weighted average F1 of CoG-BART with different proportions of SCL loss is recorded. Different $\alpha$ have a large impact on the outcomes, e.g., there exists a 2.8 points difference in F1 values between $\alpha$ equals 0.4 and 0.8 for IEMOCAP, reflecting the significant positive effect of supervised contrastive learning for this dataset. Meanwhile, different datasets have different values of $\alpha$ in obtaining the relatively best gain effect. For instance, CoG-BART performs best when $\alpha=0.2$ in MELD, while achieving the best result when $\alpha=0.4$ in IEMOCAP.

\subsection{Effect of Response Generation}
Response generation has a facilitating effect on modeling context dependence to some extent. As the two cases in Figure \ref{effect_of_response_generation} illustrate, if only the current utterance itself is considered, the expression may cause the model to misjudge the sentiment of the current utterance, while generating responses leads the model to pay more attention to contextual information, thus making correct predictions which consistent with the scenario. As for the impact of different weights of response generation loss, Table \ref{alpha_of_scl} illustrates that when fixing $\alpha$ and adjusting $\beta$, there is also a slight impact on the model's overall performance.

\subsection{Ablation Analysis}

\begin{table}[t]
\centering
\resizebox{.95\columnwidth}{!}{
\begin{tabular}{lll}
\toprule
Dataset                  & \multicolumn{1}{c}{MELD}              & \multicolumn{1}{c}{IEMOCAP}              \\
\midrule
Methods                  & \multicolumn{2}{c}{Weight-Avg-F1}                                               \\
\midrule
CoG-BART                 & \multicolumn{1}{c}{64.81}             & \multicolumn{1}{c}{66.18}               \\
\quad -Gen               & 64.26 ($\downarrow$0.55)              & 64.74 ($\downarrow$1.44)                \\
\quad -SCL loss          & 64.28 ($\downarrow$0.53)              & 64.23 ($\downarrow$1.95)                \\
\quad -Speaker           & 64.14 ($\downarrow$0.67)              & 55.41 ($\downarrow$10.77)                \\
\quad -Gen, SCL loss     & 63.57 ($\downarrow$\textbf{1.24})              & 62.90 ($\downarrow$3.28)                \\
\quad -SCL loss, Speaker & 63.72 ($\downarrow$1.09)     & 54.83 ($\downarrow$\textbf{11.35})       \\
\quad -Gen, Speaker      & 64.02 ($\downarrow$0.79)              & 54.95 ($\downarrow$11.23)                \\ 
\quad -Dialog-Trans      & 64.40 ($\downarrow$0.41)              & 64.19 ($\downarrow$1.99)                \\
\bottomrule
\end{tabular}
}
\caption{Ablation study to evaluate the impact of different components on the overall performance of the model on MELD and EmoryNLP}
\label{ablation_study}
\end{table}

To investigate the impacts of individual modules and combinations of several components on the overall effect of the model, this section conducts an ablation study on three modules in CoG-BART. As illustrated in Table \ref{ablation_study}, the selected datasets are MELD and IEMOCAP, where ``-" indicates the removal of the single method or several methods, ``Gen" denotes the auxiliary task of response generation, ``SCL loss" means supervised contrastive loss, and ``Speaker" indicates the splicing of speaker label before utterance.

From the results of MELD, removing any of the three modules makes the overall performance worse, while discarding the supervised contrastive loss and response generation has the greatest impact on the performance of CoG-BART in MELD. These indicate that supervised contrastive loss leverage label information better compared to cross-entropy loss, thus effectively distinguishing different sentiments.

Consistent results are also obtained in IEMOCAP, indicating that the improvement in model performance from these three modules transfers well across these datasets. However, the more unexpected finding was that removing the speaker's information made CoG-BART most degraded in IEMOCAP. By analyzing this  dataset, we found that it involved 302 speakers, so it may be crucial to fully model the speaker information for this dataset. It also proves the effectiveness of the simple method of splicing the speaker information directly in front of the utterance. Furthermore, removing the supervised contrastive loss alone degrades the performance by 1.95 points on IEMOCAP, indicating that supervised contrastive learning significantly impacts CoG-BART on this dataset. The results after removing Dialog-level Transformer suggest that this module improves overall performance by modelling longer contextual dependencies.

\section{Conclusion}
We propose supervised contrastive learning with response generation as an auxiliary task for BART, namely CoG-BART, for emotion recognition in conversation (ERC). First, supervised contrastive learning is introduced into the training process to distinguish similar emotions, reducing intra-class distance and increasing inter-class variance. Meanwhile, the response generation is adopted as an auxiliary task; hence, the model categorizes utterances with similar semantics but different emotions by considering the context. The results obtained on four datasets compared with the current state-of-the-art baseline methods demonstrate the proposed method's effectiveness. Furthermore, ablation studies demonstrate that supervised contrastive learning can effectively improve the model's efficacy in recognizing emotions, thus improving the overall performance. Also, response generation as an auxiliary task helps the model fully consider the context to discern the emotions of semantically similar utterances in varying contexts.

\section{Acknowledgments}
We are very grateful to the reviewers for their diligent and rigorous attitude towards our work and their valuable suggestions for improvement during the whole review process. This work was supported by the National Key Research and Development Program of China (No. 2020AAA0108702) and the National Natural Science Foundation of China (NO. 62022027).


\bibliography{CL-FOR-DER-AAAI22}
\end{document}